\documentclass{article}
\usepackage{graphicx}
\usepackage{amsmath}
\usepackage{booktabs}
\usepackage{caption}
\usepackage{subcaption}

\usepackage[accepted]{icml2026}

\icmltitlerunning{Space as Time Through Neuron Position Learning}

\begin{document}
\twocolumn[
\icmltitle{Space as Time Through Neuron Position Learning}

\begin{icmlauthorlist}
\icmlauthor{Balázs Mészáros}{sussex}
\icmlauthor{James C. Knight}{sussex}
\icmlauthor{Danyal Akarca}{imperial,cambridge}
\icmlauthor{Thomas Nowotny}{sussex}
\end{icmlauthorlist}

\icmlaffiliation{sussex}{Sussex AI, University of Sussex, Falmer, United Kingdom}
\icmlaffiliation{imperial}{Department of Electrical and Electronic Engineering, Imperial College London, London, United Kingdom
}
\icmlaffiliation{cambridge}{MRC Cognition and Brain Sciences Unit, University of Cambridge, Cambridge, United Kingdom}

\icmlcorrespondingauthor{Balázs Mészáros}{bmszros@sussex.ac.uk}

\icmlkeywords{Spiking Neural Networks, Delay Learning, Spatial Embeddings}

\vskip 0.3in
]

\printAffiliationsAndNotice{}
\begin{abstract}
    Biological neural networks exist in physical space where distance influences communication delays: a fundamental coupling between space and time absent in most artificial neural networks. While recent work has separately explored spatial embeddings and learnable synaptic delays in spiking neural networks, we unify these approaches through a novel neuron position learning algorithm where delays relate to the Euclidean distances between neurons. We derive gradients with respect to neuron positions and demonstrate that this biologically-motivated constraint acts as an inductive bias: networks trained on temporal classification tasks spontaneously self-organize into local, clustered topologies and a modular, efficiently wired structure emerges if connection costs are distance-dependent. Remarkably, we observe functional specialization aligned with spatial clustering without explicitly enforcing it. These findings lay the groundwork for networks in which space and time are intrinsically coupled, offering new avenues for mechanistic interpretability, biologically inspired modelling, and efficient implementations.
\end{abstract}
\section{Introduction}

\begin{figure*}[ht]
\begin{center}
\centerline{\includegraphics{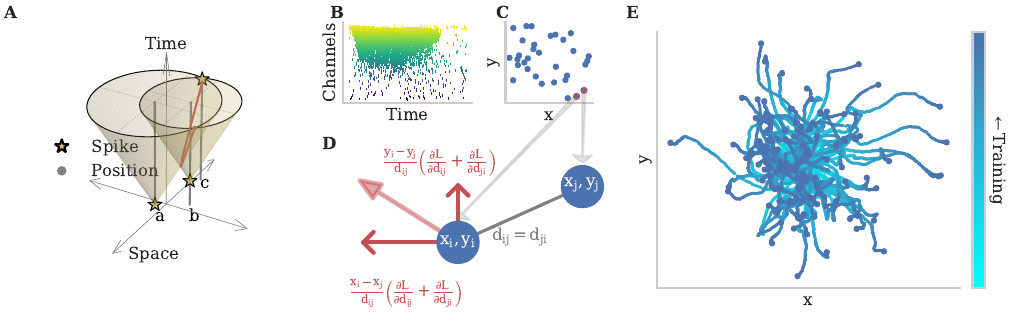}}
\caption{\textbf{(A)} A light cone represents the trajectory through spacetime of a flash of light emitted from a single point at a single instant, expanding outward in all directions over time. Similarly, in our framework, neurons' spikes (depicted as stars) “propagate” through space as they evolve in time, forming conical regions of influence in spacetime. Where these cones intersect (highlighted with a red line), the accumulated input can become strong enough to drive a neuron at that spatial location (c) to spike. \textbf{(B)} We train a single hidden layer Recurrent Spiking Neural Network on Spiking Heidelberg Digits for classification. \textbf{(C)} Compared to the standard approach, we not only train the synaptic weights, but the neuron positions as well, by defining synaptic delays as Euclidean distances. \textbf{(D)} Illustration of how individual synaptic delay gradients influence position learning updates. Neurons sum all pre- and post-synaptic delay gradients and normalise them according to the distance from the other neuron. \textbf{(E)} Training neuron positions in this framework shows the benefit of long-distance connections, compared to previous approaches, which only focused on the cost of connections. We observe that networks grow during training, with some neurons diverging more and others staying closer to each other. The lines depict the trajectory of neuron positions throughout training, and the points the final positions neurons converged on. The colours of trajectories illustrate the time course of the neuron positions throughout training.}
\label{fig:introfig}
\end{center}
\end{figure*}

Brain networks face fundamental trade-offs between metabolic costs and information processing capabilities. Networks must minimise the costs of building and maintaining connections in physical space while optimising for computational capability. This constraint shapes brain organisation across species and may explain why brains converge on similar structural solutions~\citep{achterberg2023spatially}, including sparse, clustered architectures~\citep{van2016comparative} and modular organisation~\citep{kaiser2004modelling}.

Biological neural networks exist in physical space, where distance translates directly into time through conduction delays -- a `space as time' relationship~\citep{voges2010phase, izhikevich2009polychronous, keller2024spacetime} that is fundamental to brain organisation. Recent studies have started considering the spatial embedding of biological neural networks~\citep{achterberg2023spatially,erb2026training, sheeran2024spatial,vasilache2025evolving,Tohouri2025}, focusing on the \emph{cost} of long connections, rather than also considering delays and their potential computational benefits. These studies do not investigate the relationship between the spatial embedding and time -- even though the relative location of neurons and transmission delays between them are strongly related in the brain. 

Delay learning in Spiking Neural Networks (SNNs) has recently been shown to increase the capacity of networks, i.e. to allow for compressing networks in terms of their size and their parameter precision~\citep{sun2025exploitingheterogeneousdelaysefficient}. One of the first algorithms that employed a form of delay learning used multapses with various delays between neuron pairs and a corresponding trainable weight~\citep{bohte2002error}. Since the delays were not explicitly optimised, this algorithm can be understood more as a form of structure learning where delays are fixed and the optimal values are found by optimising the connectivity weights. The relationship between network structure and delays remains an active research area.  \citet{hammouamrilearning} highlighted the benefits of delay learning in sparse networks, and~\citet{meszaros2024learning} observed that structural plasticity in networks with delays leads to better performance than explicit delay learning. Regardless, other recent works focus on gradient-based approaches~\citep{hammouamrilearning,goltz2025delgrad,sun2023learnable}, with \citet{meszaros2025efficient} introducing delay learning for recurrent connections.

The intertwinement of space and time creates natural constraints on both the structure and the dynamics of neural networks. Spatial clustering minimises costly long-range connections while creating fast communication within local regions and slower communication to distant areas. Evolution has exploited this relationship, as exemplified by the barn owl's sound localisation system, which relies on precisely tuned delays arising from spatial circuit organisation~\citep{carr1988axonal,ghosh2025spiking}.

Here, we present the natural next step: combining spatial embedding and delay learning into a single framework. Instead of learning synaptic delays directly, we define delays as the Euclidean distance between the connected neurons and derive the gradient of our loss function with respect to neuron positions. Gradient descent with respect to neuron positions effectively introduces a neural network shape learning algorithm. Within this framework, long connections, which would simply represent a high cost in previous works~\citep{achterberg2023spatially,erb2026training, sheeran2024spatial,vasilache2025evolving,Tohouri2025}, can now also be harnessed for computation. We study the various effects of the algorithm using recurrent SNNs trained on the Spiking Heidelberg Digits (SHD)~\citep{cramer2020heidelberg} classification task. We find that networks with position learning rely predominantly on local computations and form circuits associated with particular `tasks' after training. Not only do our findings have implications for our understanding of the brain, but they also suggest new ways of studying neural networks and potential directions for future neuromorphic implementations.

\section{Results}





\subsection{Position learning links structure to task}
\begin{figure*}[ht]
\begin{center}
\centerline{\includegraphics[]{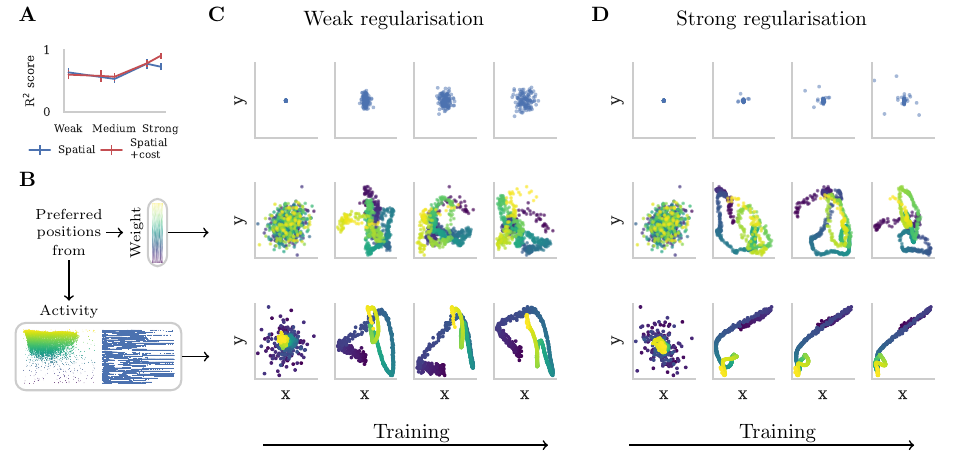}}
\caption{\textbf{(A)} $R^2$ score for ridge regression of hidden neurons' positions from their input connectivity weights. Regression has high $R^2$  for all networks, and particularly for spatial networks with strong distance-dependent cost regularisation. We depict the mean of the results over 20 random seeds, error bars represent standard deviation.\textbf{(B)} We depict the "preferred target positions" of input neurons by weighting hidden neurons' positions with the corresponding input-to-hidden weights (second row of C and D), or the correlations between input and hidden spikes (third row of C and D). \textbf{(C)} Effects of training (from left to right) in a network trained with weak L1 regularisation. First row: learnt network shapes; neurons spread out with training. Second row: preferred input target positions derived from weights. We see a clear separation of preferred target positions for the different frequency bands in the input. Last row: preferred target positions derived from activity (correlations). The positions appear to form a trajectory. \textbf{(D)} Same as C for a network with strong regularisation. Measuring the preferred target positions using the weights now also looks more like trajectories instead of clusters.}
\label{fig:prefposall}
\end{center}
\end{figure*}

\begin{figure}[ht]
\begin{center}
\centerline{\includegraphics[width=\columnwidth]{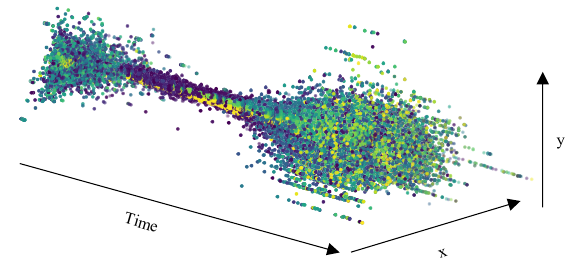}}
\caption{Preferred positions over time in an example network. Positions move towards the centre and spread out to the edges through the trial.}
\label{fig:rsta_plot}
\end{center}
\end{figure}

\begin{figure*}[ht]
\begin{center}
\centerline{\includegraphics{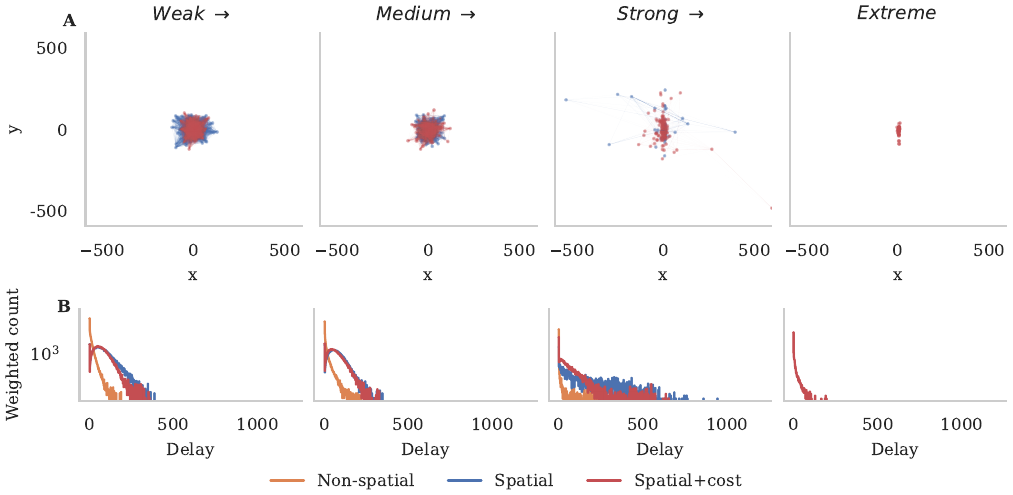}}
\caption{\textbf{(A)} Network shapes with and without distance cost and increasing regularisation strength from left to right. Line thickness corresponds to synaptic strength. Networks increase in size with stronger regularisation, until they reach a breaking point, and become small again. \textbf{(B)} Delay distributions weighted with their corresponding synaptic strength over 20 random seeds. The increase of network sizes can also be observed through the distributions.}
\label{fig:shd_reg}
\end{center}
\end{figure*}
We first introduce a novel neuron position learning algorithm, in which neurons move in space to minimise the objective function, which depends on neurons' position through synaptic delays that are defined by the Euclidean distance of neurons. 
\citet{meszaros2025efficient} were the first to introduce delay learning for recurrent connections, and, since spatial embeddings become particularly interesting with bidirectional connectivities, we derived our equations in this framework. In a two-dimensional space, we define the delay between neuron $i$ and $j$ as: 
\begin{equation}
    d_{ji} =\sqrt{(x_i-x_j)^2+(y_i-y_j)^2}, \label{eqn:Euclid}
\end{equation}
where $(x_i, y_i)$ and $(x_j, y_j)$ denote the positions of neurons $i$ and $j$.
Thus, the spike arrival time for spike $t$ at neuron $j$ becomes:
\begin{multline}
    t_{j} \equiv t +d_{ji} = t +\sqrt{(x_i-x_j)^2+(y_i-y_j)^2} ,
\end{multline}
where $i$ is the neuron that fired at time $t$, which arrived at neuron $j$ at time $t_j$. Similar to the original algorithm~\citep{wunderlich2021event} and its extensions~\citep{meszaros2025efficient}, we also work with Leaky Integrate-and-Fire (LIF) neurons, where each neuron has an input current $I$, and a membrane voltage $V$. 

The gradient of the loss $\cal L$ with respect to the coordinate $x$ of neuron $i$ can be derived from the gradient with respect to the delay ($d_{ji}$), using the chain rule:
\begin{align}
    \frac{d\mathcal{L}}{dx_{i}} = \sum_j \left(\frac{d\mathcal{L}}{d d_{ji}}\frac{dd_{ji}}{dx_{i}} + \frac{d\mathcal{L}}{d d_{ij}}\frac{dd_{ij}}{dx_{i}} \right)
\end{align}
From \citet{meszaros2025efficient}, equation (6), we know
\begin{align}
    \frac{d\mathcal{L}}{d d_{ji}} = - w_{ji} \sum_{ t_k \in \,\parbox{0.65cm}{\scriptsize spikes from\ $i$}} (\lambda_I-\lambda_V)_j\Big|_{t_k+d_{ji}} \label{eq:dldd}
\end{align}
where $w_{ij}$ is the synaptic weight; and $\lambda_I$ and $\lambda_V$ denote the backwards sensitivities corresponding to the input current and membrane voltage, respectively.
Next, by taking the derivative of equation (\ref{eqn:Euclid}), we get
\begin{align}
    \frac{dd_{ji}}{dx_{i}} = \frac{x_{i}-x_{j}}{d_{ji}} = \frac{dd_{ij}}{dx_{i}}.
\end{align}
Taken together with the expression for $\frac{d\mathcal{L}}{d d_{ij}}$, obtained by rewriting (\ref{eq:dldd}), we obtain:
\begin{multline}
    \frac{d\mathcal{L}}{dx_{i}}  = - \sum_j \frac{x_{i}-x_{j}}{d_{ji}} \Bigg(\sum_{t_k \in \,\parbox{0.65cm}{\scriptsize spikes\\ from\ $i$}} w_{ji}\left.(\lambda_I-\lambda_V)_j\right|_{t_k+d_{ji}} \\
    + \sum_{t_k \in \,\parbox{0.66cm}{\scriptsize spikes\\ from\ $j$}} w_{ij}\left.(\lambda_I-\lambda_V)_i\right|_{t_k+d_{ij}}\Bigg). \label{eq:x_grad}
\end{multline}

We investigate our algorithm by training SNNs with learnable positions on the SHD classification task~\citep{cramer2020heidelberg}. By learning positions instead of synaptic delays, we reduce the number of free parameters from $n^2$ to $n\times d$, where $n$ is the number of neurons, and $d$ is the dimensionality of the space neurons are embedded in. Furthermore, this set-up yields delay distributions which inherit the following properties from the Euclidean distance:
\begin{itemize}
    \item Zero diagonals: Spikes arriving through self connections are immediate
    \item Symmetry: The delay from neuron $i$ to neuron $j$ is the same as from $j$ to $i$ 
    \item Triangle inequality: The `quickest' path to another neuron is always the direct path
\end{itemize}
To ascertain that position learning is useful, we first compared the classification accuracy of position learning networks and standard networks with and without delays. Recurrent delay learning through positions performs similarly to non-spatial networks with fewer trainable parameters (supplementary figure \ref{fig:shd_acc_space} in the Appendix). For the remainder of this paper, we focus on analysing 2D networks with a hidden layer size of 128 since we observed no fundamental differences between network sizes and embedding dimensions.

In the following, we study unconstrained synaptic delay learning (Non-spatial) and position learning (Spatial) networks with L1 regularisation on the hidden weights, as well as a spatially weighted L1 regularisation (Spatial+cost) in which each synaptic weight is penalised proportionally to the distance between neurons \citep{achterberg2023spatially}. In the following, we study unconstrained synaptic delay learning (Non-spatial) and position learning (Spatial) networks with L1 regularisation on the hidden weights, as well as a spatially weighted L1 regularisation (Spatial+cost), in which each synaptic weight is penalised proportionally to the distance between neurons \citep{achterberg2023spatially}. In the latter case, we use the normal L1 penalty and apply its gradient, scaled by distance,  to the hidden weight updates. Note that this implies that L1 regularisation is only applied to weights, and not to positions. This distance-dependent penalty can be interpreted as a resource constraint that discourages both excessive connectivity and long-range connections, reflecting limitations on wiring and metabolic cost.
Since the distances between neurons change dynamically during training, when calculating the spatially weighted L1 regularisation, we normalise the distances by dividing by the mean. This ensures that the overall strength of the regularisation term remains comparable to standard L1 regularisation and does not dominate the loss function solely due to changes in network geometry.

\subsection{Nearby neurons become functionally connected}

First, we focus on the relationship between the computational role of individual neurons and their spatial position. We looked at the relationship between neuron positions in the hidden layer and the inputs they preferentially respond to after training. First, we predicted the $x$, $y$ positions of hidden neurons from their corresponding 700-dimensional input to hidden weight vectors using ridge regression. We found that -- based on the observed $R^2$ scores shown in Figure~\ref{fig:prefposall}A -- that we can do this accurately, suggesting that neurons' input patterns are strongly correlated to their position. We observe an increase in $R^2$ scores as regularisation strengthens.

To get an understanding of \emph{where} computations are happening, we also looked at the `centre of mass' of each input neuron's projections, by taking the mean of all hidden neuron positions weighted by the corresponding input to hidden synaptic strength (Figure~\ref{fig:prefposall}B, equation~\ref{eq:prefpos} in the Appendix). As the examples in Figure~\ref{fig:prefposall}C show, after optimising neuron positions, input neurons representing similar audio frequencies appear to project to specific locations. We also observe that, as regularisation strength increases, preferred target locations for input neurons representing similar frequencies not only become more localised, but they even organise in to structures that look like trajectories (Figure~\ref{fig:prefposall}D).

Next, we measured the influence of input neurons on hidden layer activity by analyzing evoked hidden layer spikes. We measured the average spiking activity (across input samples) of each hidden neuron in the 5 timesteps following each input spike (for definition see equation~\ref{eq:prefpos2} in the Appendix).
We then took the average of this 3-dimensional tensor (input neuron $\times$ hidden neuron $\times$ 5) over the 5 timesteps. This process again results in a matrix with the same shape as the input-to-hidden weight matrix.
Therefore, we can again calculate the `centre of mass' of each hidden neuron's influence based on these values. Figure \ref{fig:prefposall}C and D show examples of this process, and depicting preferred target positions in this manner shows the formation of clear trajectories, regardless of regularisation strength. 

Importantly, both clustered and trajectory-like patterns indicate non-random spatial organisation. While clustering reflects the localisation of inputs with similar frequencies to nearby hidden neurons, the trajectory-like structures show a continuous spatial organisation of input frequency in the hidden layer.

\begin{figure*}[ht]
\begin{center}
\centerline{\includegraphics{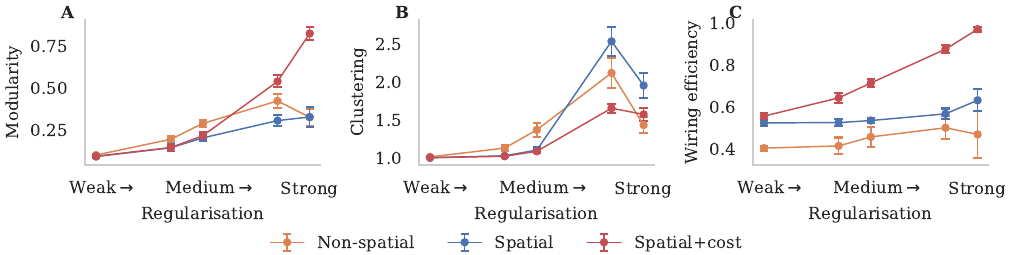}}
\caption{Regularisation effects on topology. We depict the mean of the results over 20 random seeds, error bars represent standard deviation. \textbf{(A)} Weighted modularity. `Spatial+cost' network shows a monotonic increase, while the modularity of other networks seems to peak with weaker regularisation. \textbf{(B)} Weighted clustering coefficient. Resource constraints increases clustering in all cases, but plain L1 regularisation has a stronger effect. Spatial networks without the distance cost achieve particularly high clustering. \textbf{(C)} Weighted wiring efficiency. Following our intuition, efficiency is particularly high with a distance cost.}
\label{fig:shd_reg2}
\end{center}
\end{figure*}

Instead of taking the average over time of the previously measured `sensitivities', we can also measure the average sensitivity at each timestep and observe where computation is happening for each input neuron. As shown in Figure~\ref{fig:rsta_plot}, early in the trial, preferred positions are concentrated towards the spatial centre of the network where neurons are closer to each other and therefore primarily connected via short synaptic delays. Later on, preferred positions spread outward towards more peripheral regions, the interactions between which rely on longer delays due to the increased distances.
This temporal progression suggests a shift from fast, short-delay interactions supporting rapid feature extraction early in the trial to longer-delay interactions later on, enabling information to be maintained and integrated over extended timescales. Such behaviour is consistent with the notion of temporal hierarchies in neural computation \citep{moro2024role}.

\subsection{Relative importance of short and long delays under constraints}

As expected, increasing L1 weight regularisation strength results in lower accuracy and higher sparsity (see Figure \ref{fig:shd_reg_acc} in the appendix). Figure~\ref{fig:shd_reg}A visualises the shapes of example networks with increasing regularisation strengths both with and without the distance cost. As regularisation strength increases, both the network with and without spatial cost increase in size (with the long distance synapses being stronger in the absence of the distance cost). However, at extreme regularisation strength, we reach a breaking point, where networks become very small again. Note that, for this regularisation strength, the non-spatial network and the spatial network without the distance cost converge on solutions that do not rely on the recurrent connections, i.e. all recurrent connections can be pruned without accuracy loss. We speculate that as the networks have fewer weights (with increased weight regularisation), they have to depend on longer and longer delays. However, while long delays can be introduced on individual connections in the non-spatial architecture, in spatial networks, this can only be achieved by moving neurons further apart, which increases \emph{all} delays associated with these neurons. Thus, extreme resource constraints show the importance of short delays. Observing the synaptic strength-weighted delay distributions (figure \ref{fig:shd_reg}B) reveals that across architectures and regularisation strengths, delays seem to be concentrated in the short range.

\subsection{Learnt topology is constraint-dependent}

Next, we studied the topology of the learnt network under increasingly strong regularisation. In Figure \ref{fig:shd_reg2}A, we can observe that, as the regularisation strength increases, the network with the additional distance cost achieves higher and higher modularity (for definition see equation (\ref{eq:q_metric}) in the Appendix). In the other networks, while modularity increases to some extent, it reaches its maximum observed value with weaker regularisation than the maximum studied strength.

We also calculated a weighted clustering metric from our connectivity matrices, using a shuffled weight matrix as the null-model (for definition see equation (\ref{eq:clustering}) in the Appendix). Interestingly, we see a rapid increase for the non-spatial and spatial network, particularly compared to the network with distance cost. However, again, stronger regularisation does not increase the value further. In networks with the distance cost we observe a similar trend, but with lower values.

Although most connections in biological neural networks are short -- reflecting spatial and energetic constraints -- such networks typically contain a small but important set of long-distance connections. Thus, the wiring cost is higher than a network rewired for minimum cost, but much lower than a network rewired for maximum cost \citep{kaiser2006nonoptimal}. Thus, in our networks, we measure how the wiring cost relates to the minimum and maximum possible wiring cost (see equation (\ref{eq:efficiency}) in the Appendix for definition). We observe that our spatial networks are more efficiently wired than non-spatial architectures and show higher efficiency at higher regularisation strengths. Intuitively, strong distance-scaled regularisation yields particularly efficient wiring. 

\subsection{Unconstrained position learning converges on more local solutions}

\begin{figure*}[ht]
\begin{center}
\centerline{\includegraphics{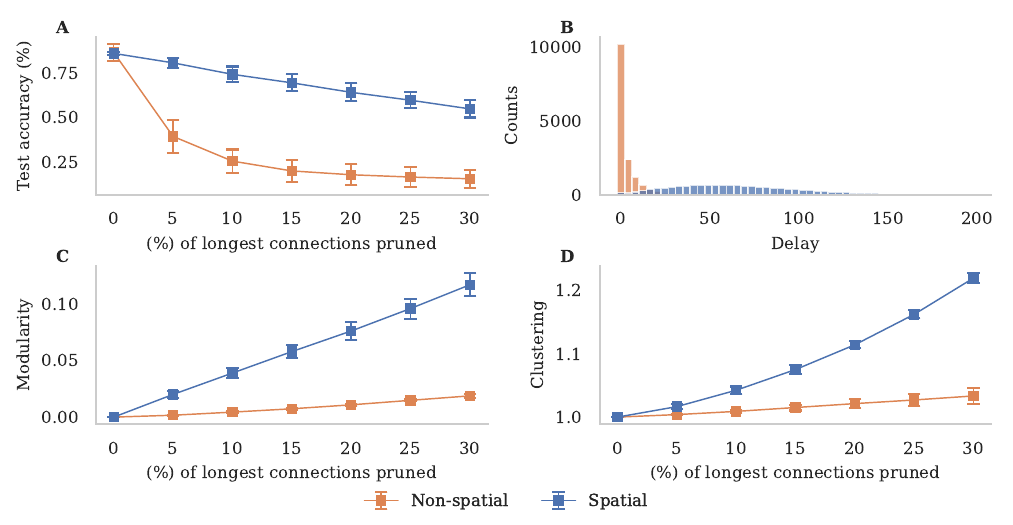}}
\caption{Results of pruning longest connections. We depict the mean of the results over 20 random seeds, error bars represent standard deviation. \textbf{(A)} Pruning effects on classification accuracy on SHD. Pruning is less harmful for the position learning network. \textbf{(B)} Delay distributions post training. For the non-spatial network, most delays are still zero after training, but we also end up with a few very long delays that are not visible since they are significantly outnumbered by the zero delays. For the spatial network delays become spread out in the lower range. \textbf{(C)} Delay length based pruning effects on modularity. Position learning network achieves higher modularity. \textbf{(D)} delay length-based pruning effects on clustering. Position learning network achieves higher clustering.}
\label{fig:shd_prune_all}
\end{center}
\end{figure*}
Finally, we also study networks with no regularisation.
This is, of course, not biologically realistic as connecting neurons in the brain always has a `cost' (particularly if they are distant). However, we were interested in what solutions the network finds without constraints and, whether networks still converge on solutions where certain synapses are more `important' than others. We found that `classic' weight magnitude-based pruning and random pruning had very similar effects on both non-spatially embedded networks and those where positions were learnt. Taking inspiration from biology and emphasising short connections, we pruned away the top $x\%$ (ranging from $5\%$ to $30\%$) longest connections. The learnt delay distributions of non-spatial and spatial networks shown in figure \ref{fig:shd_prune_all}A seem to hint that spatial networks converge on solutions which are more reliant on short local connections. Given that this was not enforced through any explicit regularisation, this is a surprising finding, particularly since long delays tend to be beneficial for temporal tasks~\citep{sun2025exploitingheterogeneousdelaysefficient}. These findings not only align with findings from neuroscience~\citep{swadlow1985physiological} but also have implementation benefits, as, in digital neuromorphic systems, delays are usually implemented through costly dynamic memory buffers whose size has a strict upper bound~\citep{davies2021advancing}. Therefore, shortening delays can be beneficial. Given that random and magnitude-based pruning did not show such differences, seemingly spatial networks are not simply more robust but are more `local'. We also tested robustness through neuron ablation studies and adding noise to delays, neither of which resulted in a significant difference between spatial and non-spatial architectures.

Since delay-based pruning has less impact on the performance of spatial architectures, we were curious to see how the network topology changes after pruning. Therefore, we recalculate modularity and clustering in our pruned networks (since the connectivity is now sparse, we disregard the weights and measure these metrics on the binary connectivity matrices using equations (\ref{eq:q_metric}) and (\ref{eq:clustering}) in the Appendix).
Figure ~\ref{fig:shd_prune_all}C-D shows the effect of delay-based pruning on modularity and clustering. As a baseline, we calculated the modularity of a random matrix, and found that spatial architectures show a clear increase in modularity with pruning, while non-spatial architectures show a similar trend as pruning a random matrix. Similarly, clustering increases more rapidly in spatial networks (with the same connectivity ratio). After pruning, we observe higher accuracy, modularity and clustering in spatial architectures, without explicitly enforcing any of these structures to emerge.

\section{Discussion}


Here, we have introduced a method for studying the intertwinement of space and time in spiking neural networks.
While neuron position learning has been explored before, prior studies have mainly emphasised parameter efficiency~\citep{erb2026training,Tohouri2025,landsmeer2025spatial} or performance improvements~\citep{vasilache2025evolving} on machine learning benchmarks. Here, we instead focus on how the structure of learnt networks changes when space and time are intrinsically connected through delays. 

\citet{achterberg2023spatially} investigated networks with fixed neuron positions. If positions were instead allowed to be optimised, since distance acts only as a cost, all neurons would likely converge toward the centre, meaning the benefits of modularity and clustered structure would vanish. Once proximity carries no penalty, the network is free to adopt any connectivity pattern needed to solve the task. In contrast, by introducing a benefit of distance, our framework prevents this collapse and encourages spatially distributed structures. Similarly, in biological neural networks, most connections are short-range, yet the overall average connection length remains higher than the physical minimum~\citep{bassett2010efficient}. During development, neuronal migration~\citep{de2006ontogeny} contributes to functional circuit formation. However, biological constraints such as each neuron occupying a physical volume (volume exclusion) prevents neurons from being too close to one another. In our model, neurons can theoretically be arbitrarily close, provided they are not coincident. Introducing an explicit repulsive force between two neurons would be straightforward to formalise but challenging to extend to more complex circuits -- offering another interesting direction for future work. Furthermore, now we even have studies that track neurons' position changes over time \citep{10.7554/eLife.85300}, allowing for direct comparisons between the presented algorithm and empirical data.

We find that the positions of neurons in the hidden layer correlate with input to hidden connectivity patterns. From a mechanistic interpretability perspective, these correlations point to concrete computational mechanisms. In our simple classification setting, the precise functional roles of clusters of nearby neurons remain unclear. However, this is likely due to the limited temporal structure of the task rather than a lack of internal specialisation. Long delays between modules specialised for different subtasks may be advantageous when temporal alignment across these modules is required for solving a more complex task. Therefore, extending this analysis to modular tasks~\citep{bena2025dynamics} with an added temporal structure, to multimodal settings, or multitask problems~\citep{vafidisdisentangling} provide a promising avenue for extending this mechanistic analysis. In our examples, we focus on the representation of inputs in the hidden layer. Of course, it would also be interesting to see if neurons (or even groups of neurons) specialise on particular inputs, but since we are compressing from the input to the hidden layer (700 input neurons to 128 hidden neurons), we would not expect specialisation like that to emerge. As mentioned before, we study small networks to make sure that delays become useful for the task. We could also constrain the networks by only allowing hidden neurons to spike once or by turning to tasks with low-dimensional but complex inputs, meaning that recurrent delays remain crucial even with a larger hidden layer.

Depending on the constraints imposed during training, our networks consistently converge to structured spatial organisations which have distinct practical implications for neuromorphic hardware. Networks trained with position learning converge to solutions that exploit synaptic delays while keeping them short, which is desirable for digital neuromorphic implementations~\citep{davies2021advancing}. Imposing L1 regularisation scaled by interneuron distance leads to highly modular architectures. Such modular designs are particularly well suited to resource-constrained autonomous agents, which naturally favour modular organisation~\citep{bartolozzi2022embodied}. Interestingly, we observe that while distance cost leads to clustered architectures, imposing pure L1 regularisation achieves even higher values. This is directly relevant for recent neuromorphic hardware designs, such as Mosaic~\citep{dalgaty2024mosaic}, which explicitly relies on clustered connectivity. \citet{weber2025hardware} demonstrated that the structural plasticity algorithm DEEP R \citep{bellec2018deep} can be extended to account for system constraints (e.g., small-world topology). Inspired by this, introducing distance awareness with position learning to DEEP R could yield interesting network geometries, as neuron positions would no longer need to be constrained to a hypersphere, but merely comply with system constraints. Furthermore, our results seem to imply that using distance-dependent delays, we could find architectures that are inherently more clustered, and thus could make the best use of hardware which requires clustered connectivity.

In this work, we deliberately focus on small networks in order to enable detailed analysis and clear attribution of observed effects. However, this choice does not reflect an inherent limitation of SNNs, which have been successfully trained at large-scales~\citep{zhu2024spikegpt}. Scaling is particularly interesting in the present context, as spatially local connectivity and distance-dependent delays are key ingredients for the formation of travelling waves~\citep{davis2021spontaneous}. \citet{keller2024spacetime} outlines the importance of spacetime neural representation both for theoretical neuroscience and improved generalisation and long-term working memory in AI. Travelling waves, in particular, have been shown to facilitate global information integration~\citep{jacobs2025traveling,keller2023neural,liboni2025image}.

\section*{Software and Data}
The data underlying our results will be available at \url{https://doi.org/10.25377/sussex.30455591}. All experiments were carried out using the GeNN 5.1.0 and mlGeNN 2.3.0. The version used for this study is available at \url{https://github.com/mbalazs98/ml_genn/tree/position_learning}. The code to train and evaluate the models described in this work are available at \url{https://github.com/mbalazs98/spatialdelays}.

\section*{Acknowledgments}
This work was funded by the EPSRC (grants EP/V052241/1 and EP/S030964/1) and the EU (grant no. 945539). Additionally, BM was funded by a Leverhulme Trust studentship and by The Alan Turing Institute, and DA was funded by an Imperial College Research Fellowship, Schmidt Sciences Fellowship and Templeton World Charity Foundation, Inc (funder DOI 501100011730) under the grant TWCF-2022-30510. Compute time was provided through Gauss Centre for Supercomputing (application numbers 21018, 30182 and 61883) and EPSRC (grant number EP/T022205/1) and local GPU hardware was provided by an NVIDIA hardware grant award.

\section*{Impact Statement}

This paper presents work aiming to advance the field of Machine
Learning. If developed in further applied research, there may be many potential societal consequences of our work, none of
which we feel must be specifically highlighted here.

\bibliography{sample}
\bibliographystyle{icml2026}

\newpage
\appendix
\section{Comparing accuracy of networks}
\begin{figure}[ht]
\begin{center}
\centerline{\includegraphics[width=\columnwidth]{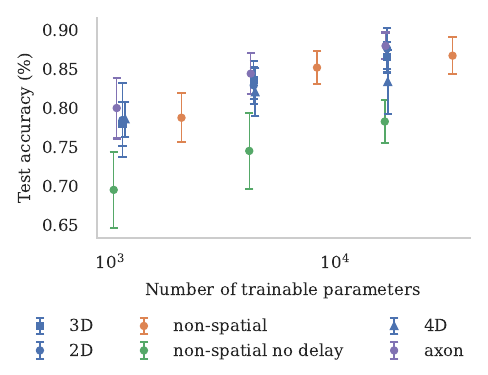}}
\caption{Network comparisons. We depict the mean of the results over 20 random seeds, error bars represent standard deviation.}
\label{fig:shd_acc_space}
\end{center}
\end{figure}

To ascertain that position learning is useful, we first compare position learning networks of hidden layer sizes between 32 and 128 and embeddings of 2, 3 and 4 dimensions, against networks with non-spatially constrained synaptic, and axonal delays, and networks without any delays. Axonal delays mean that all connections originating from the same neuron have an identical delay, i.e. the delay matrix becomes a delay vector. We could have additionally compared against networks with fixed random delays, but it raises the question of how to initialise these delays. In recurrent connections, this is a particularly complex and understudied question. We illustrate the comparisons of the networks in figure \ref{fig:shd_acc_space}.

\section{Preferred positions}

We calculate the preferred position of each input neuron $i$ in the $x$ dimension with the following equation:
\begin{equation}
    {x^{\mathrm{pref}}_i}
    =
    \frac{\sum_j { w_{ij}}\,{ x_j}}
         {\sum_j { w_{ij}}}
\label{eq:prefpos}
\end{equation}
where $w_{ij}$ is the learnt synaptic weight between input neuron $i$ and hidden neuron $j$. Instead of using the learn weights, we can also calculate `sensitivities' ($\tilde{w}_{ij}(t)$) , derived from neuron activities:

\begin{equation}
{
\tilde{w}_{ij}(t)=\bf{1}_{\{t={t_k^i} \; \land \;|t-{t_n^j}|<\Delta\}},
}
\label{eq:prefpos2}
\end{equation}
where $t_k^i$ denotes the time of the k-th spike of input neuron $i$, and $t_n^j$ denotes the time of the n-th spike of hidden neuron $j$.

 With a fixed number of trainable parameters, we observe that networks with delays always outperform networks without them. As network size (i.e. number of hidden neurons) increases, the performance gaps decrease. This is not surprising, as all architectures have recurrent connections, which, similar to delays, can serve as a temporal memory. As the network gets more and more neurons to store information in, delays become less and less useful. Spatially embedded networks perform similarly to networks with unconstrained delay learning. Axonal delays, however, perform at least as well as the synaptic delays derived from the Euclidean distances. Note, that the error bars are large because we study small networks, to make sure that there is a clear benefit of delays.
\section{Regularisation}\label{regularisation}

For this work, the original EventProp backward dynamics~\citep{wunderlich2021event,nowotny2025loss} do not change. The weight update rule with delays~\citep{meszaros2025efficient} is defined as
\begin{equation}
    \frac{{\rm d}\mathcal{L}}{{\rm d}w_{ji}} =-\tau_s \sum_{t_kfromi}\lambda_{I,j}\Big|_{t_k +d_{ji}},
\end{equation}
where $\mathcal{L}$ is the loss, $w_{ji}$ is the synaptic weight from neuron $i$ to $j$, $\tau_s$ is the synaptic time constant, $t_k$ is the k-th spike, $\lambda_I$ is the input current backward dynamic for neuron $j$, and $d_{ji}$ is the delay between neuron $j$ and $i$. If we apply an L1 regularisation term, the update rule becomes
\begin{equation}
    \frac{{\rm d}\mathcal{L}}{{\rm d}w_{ji}} =-\tau_s \sum_{t_kfromi}\lambda_{I,j}\Big|_{t_k +d_{ji}}+\Lambda_1 sign(w_{ji}),
\end{equation}
where $\Lambda_1$ is the regularisation strength. If we introduce the mean scaled distance cost as well, the equation is
\begin{equation}
    \frac{{\rm d}\mathcal{L}}{{\rm d}w_{ji}} =-\tau_s \sum_{t_kfromi}\lambda_{I,j}\Big|_{t_k +d_{ji}}+\Lambda_1 sign(w_{ji})d_{ji}/\bar{D},
\end{equation}
where $\bar{D}$ is the mean of the delay matrix. The regularisation term was omitted from the neuron position updates (i.e. the gradient of the regularisation term was directly added to weight gradients and not to the loss function), so that only the topology of the network was effected, and not the geometry.

\begin{figure}[ht]
\begin{center}
\centerline{\includegraphics[width=\columnwidth]{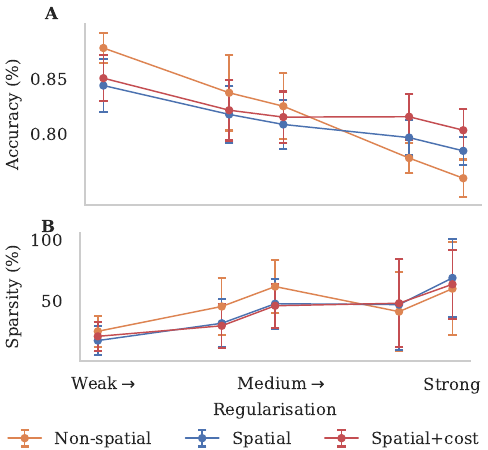}}
\caption{We depict the mean of results over 20 random seeds, error bars represent standard deviation. \textbf{(A)} Classification accuracy over increasing regularisation strength. \textbf{(B)} Connection sparsity over increasing regularisation strength.}
\label{fig:shd_reg_acc}
\end{center}
\end{figure}
\section{Modularity}
We define the modularity statistic Q as
\begin{equation}
    Q = \frac{1}{l} \sum_{i,j\in N}\left(a_{i,j}-\frac{k_ik_j}{l}\right)\delta_{m_im_j}, \label{eq:q_metric}
\end{equation}
where $l$ is the total number of synapses (or the the sum of all synapse strengths in the weighted case), $N$ is the total number of neurons, $a_{ij}$ is the connection status between neuron $i$ and $j$ (or the synapse strength in the weighted case), $k_i$ is the number of synapses of neuron $i$, and $m_i$ is the module containing neuron $i$.
\section{Small-worldness}
We define clustering as
\begin{equation}
    C = \frac{1}{N} \sum c_{i,j},\label{eq:clustering}
\end{equation}
where $c_{i,j}$ is the number of closed triangular motifs including node $i$. We normalise this by the clustering of a null-model of the same density:
\begin{equation}
    \Gamma = \frac{C}{C_{rand}}.\label{eq:clusteringnorm}
\end{equation}
The weighted clustering is defined as:
\begin{equation}
    C_w = \frac{1}{k_i(k_i-1)} \sum_{j,k}(w_{ij}w_{jk}w_{ik})^{1/3},\label{eq:wclustering}
\end{equation}
normalised by a null-model which is constructed by random permutations of the weight matrix. Figure \ref{fig:shd_reg} shows increased weighted clustering with increased regularisation stength, and figure \ref{fig:shd_prune_all} shows increased clustering with pruning the spatial network. In the binary case, we define the path length as:
\begin{equation}
    L = \frac{1}{N} \sum l_{i,j},\label{eq:path}
\end{equation}
where $l_{i,j}$ is the shortest path between nodes $i$ and $j$, and again, we normalise with a null-model of the same density:
\begin{equation}
    \Lambda = \frac{L}{L_{rand}}.\label{eq:pathnorm}
\end{equation}
The definition of path length in the weighted case is:
\begin{equation}
    L_w = \frac{1}{N(N-1)} \sum_{i \neq j} \frac{1}{w_{i,j}},\label{eq:path}
\end{equation}
normalised again by the average path-length of random permutations of the weight matrix. With these two equations, we can define small-worldness as:
\begin{equation}
    \sigma = \frac{\Gamma}{\Lambda}.\label{eq:smallworld}
\end{equation}

By definition, we consider a network small-world if $\sigma > 1, \Gamma>1$ and $\Lambda \approx 1$. We observed high values of $\Gamma$ in our experiments, thus if $\Lambda \approx 1$, we have small-world networks.
\begin{figure}[ht]
\begin{center}
\centerline{\includegraphics[width=\columnwidth]{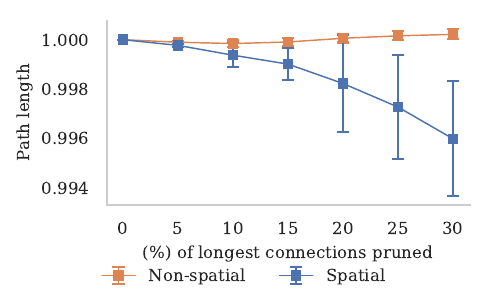}}
\caption{Path length after pruning longest delays. We depict the mean of the results over 20 random seeds, error bars represent standard deviation.}
\label{fig:binary_path_length}
\end{center}
\end{figure}
We observe in figure \ref{fig:binary_path_length} that for the non-regularised, pruned networks the path length insignificantly decreases from the initial value of $1$, thus, we conclude pruning the spatial network yields leads to small-world topologies.
\begin{figure}[ht]
\begin{center}
\centerline{\includegraphics[width=\columnwidth]{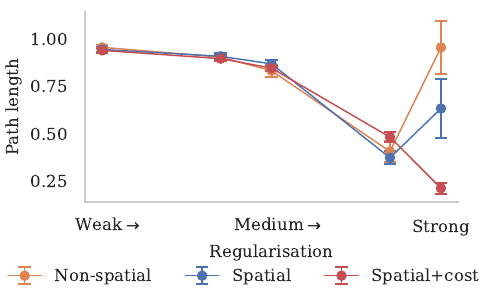}}
\caption{Weighted path length with various regularisation strengths. We depict the mean of the results over 20 random seeds, error bars represent standard deviation.}
\label{fig:weighted_path_length}
\end{center}
\end{figure}

In contrast, for the regularised networks (figure \ref{fig:weighted_path_length}) we can see that the weighted path length significantly decreases with stronger regularisation. This might be due to the network converging on solutions with multiple smaller populations of neurons operating separately in the hidden layer. Thus, these networks are not small-world.

\section{Wiring efficiency}

We define the wiring efficiency by

\begin{equation}
    \text{wire} = \langle W,D\rangle_F,\label{eq:efficiency}
\end{equation}
and normalise these values between $0$ and $1$, by calculating the maximum and minimum possible wiring, with the learnt network geometry. By sorting the weights in an ascending order, and the delays in a descending order and again calculating the product, we get the minimum possible wiring ($\text{wire}_\text{min}$), and by ordering both in a descending (or ascending) order we get the maximum possible wiring ($\text{wire}_\text{max}$.We then normalise the efficiency with the following equation:
\begin{equation}
    \text{wire}_\text{norm} = 1-\frac{\text{wire}-\text{wire}_\text{min}}{\text{wire}_\text{max}-\text{wire}_\text{min}}\label{eq:efficiency_norm}
\end{equation}
\section{Entropy}
Shannon entropy quantifies the amount of unpredictability. A highly predictable system has low entropy and will exhibit a degree of clustering around particular weight values. An unpredictable system has high entropy and will be more uniform in its distribution. While the previously introduced modularity metric Q shows the extent to which the network can be partitioned into subcommunities, it does not measure the degree of uncertainty of the weight distribution. The Shannon entropy of the weight matrix $W$ is defined as
\begin{equation}
    H(W)=-
    \frac{1}{N}\sum_{i=1}^N\frac{w_{ij}}{\sum_{i=1}^Nw_{ij}}log_2\Big(\frac{w_{ij}}{\sum_{i=1}^Nw_{ij}}\Big),
\end{equation}

\begin{figure}[ht]
\begin{center}
\centerline{\includegraphics[width=\columnwidth]{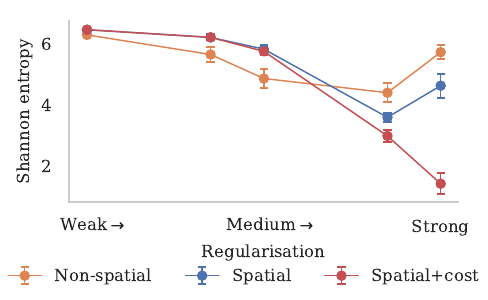}}
\caption{Shannon entropy over various regularisation strengths. We depict the mean of the results over 20 random seeds, error bars represent standard deviation.}
\label{fig:shd_ent}
\end{center}
\end{figure}

In Figure \ref{fig:shd_ent} we observe, that position learning networks with distance cost achieve the lowest entropy, in contrast to others which increase in entropy when regularisation is strong.

In this work, we are interested in the effect of space on network structure, and thus only introduced the distance cost to the regularisation term. \citet{achterberg2023spatially} and \citet{sheeran2024spatial} also introduced a communicability cost, which we did not employ here. However, we can still measure the communicability entropy, similarly to \citet{sheeran2024spatial}. 
We define the communicability matrix as
\begin{equation}
    C = e ^{-\frac{1}{2}}We ^{-\frac{1}{2}},
\end{equation}
and we can calculate the communicability entropy as
\begin{equation}
    H(W)=-
    \frac{1}{N}\sum_{i=1}^N\frac{C_{ij}}{\sum_{i=1}^NC_{ij}}log_2\Big(\frac{C_{ij}}{\sum_{i=1}^NC_{ij}}\Big).
\end{equation}

\begin{figure}[ht]
\begin{center}
\centerline{\includegraphics[width=\columnwidth]{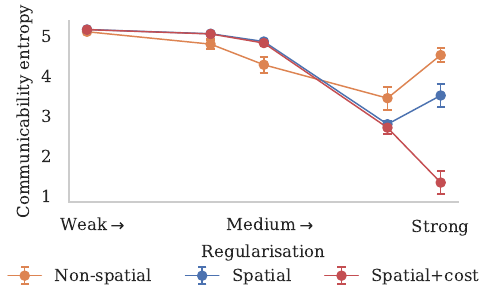}}
\caption{Communicability entropy over various regularisation strengths. We depict the mean of the results over 20 random seeds, error bars represent standard deviation.}
\label{fig:shd_cent}
\end{center}
\end{figure}
In figure \ref{fig:shd_cent}, we can observe that networks with the introduced distance cost yield a lower entropy. In contrast to the Shannon entropy, the spatial network without the distance cost shows a decrease in communicability entropy as the regularisation strength increases. These architectures yield a small number of highly communicable connections, with an increasingly ordered topology.

\section{Implementation}
For training our models we used the mlGeNN library \citep{knight2023easy}. We implemented support for weight regularisation, the position gradient update function, and axonal delays. For our graph measures we used the Brain Connectivity Toolbox \citep{rubinov2010complex}.
\end{document}